\documentclass[10pt,twocolumn,letterpaper]{article}

\usepackage{cvpr}
\usepackage{times}
\usepackage{epsfig}
\usepackage{graphicx}
\usepackage{amsmath}
\usepackage{amssymb}
\usepackage{tabularx}
\usepackage{multirow}

\newcommand*\rot{\rotatebox{90}}
\usepackage[pagebackref=true,breaklinks=true,letterpaper=true,colorlinks,bookmarks=false]{hyperref}

\cvprfinalcopy 

\ifcvprfinal\fi
\begin{document}

\title{Split-Merge Pooling}

\author{{Omid Hosseini Jafari \hspace{2pt} and \hspace{2pt} Carsten Rother}\\
Visual Learning Lab\\
Heidelberg University (HCI/IWR)\\
{\tt\small http://vislearn.de}
}

\maketitle

\begin{abstract}
There are a variety of approaches to obtain a vast receptive field with convolutional neural networks (CNNs), such as pooling or striding convolutions. Most of these approaches were initially designed for image classification and later adapted to dense prediction tasks, such as semantic segmentation. However, the major drawback of this adaptation is the loss of spatial information. Even the popular dilated convolution approach, which in theory is able to operate with full spatial resolution, needs to subsample features for large image sizes in order to make the training and inference tractable. In this work, we introduce Split-Merge pooling to fully preserve the spatial information without any subsampling. By applying Split-Merge pooling to deep networks, we achieve, at the same time, a very large receptive field. We evaluate our approach for dense semantic segmentation of large image sizes taken from the Cityscapes and GTA-5 datasets. We demonstrate that by replacing max-pooling and striding convolutions with our split-merge pooling, we are able to improve the accuracy of different variations of ResNet significantly. 
\end{abstract}

\section{Introduction}
Convolutional neural networks (CNNs) are the method of choice for image classification \cite{RanzatoHBL07, Alexnet, VGG, He_2016_CVPR, imagenet}. CNNs capture multi-scale contextual information in an image via subsampling the intermediate features through the network layers \cite{RanzatoHBL07,SchererMB10}. This approach is successful due to the expansion of the receptive fields, which are large enough to capture the context of the image for classification. 

In this work, we consider dense prediction tasks, which are popular in computer vision \cite{taskonomy2018}. One desideratum of dense prediction tasks is to have pixel-accurate predictions, for example in semantic segmentation \cite{He_2016_CVPR,long-shelhamer-fcn-2015,Noh_2015_ICCV,U-net,Hariharan_2015_CVPR} or depth estimation \cite{eigen-nips14, eigen-iccv15}. Even small inaccuracies, such as missing a small object lying on the street, may lead to an accident of an autonomous vehicle.

\begin{figure*}[!htb]
    \centering
    \includegraphics[width=0.95\linewidth]{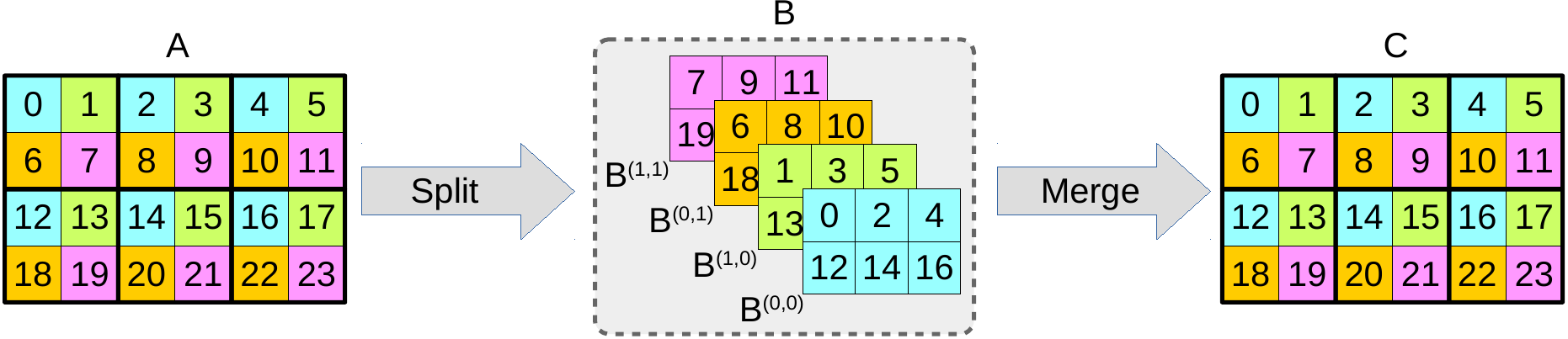}
    \caption{\textbf{Split and Merge Pooling.} The illustration of split and  merge pooling layers with a window size of $2\times 2$. The advantage of splitting the input into batches is to make it possible to process each part of the input (\ie each split batch) independently. In this example, after splitting $A$ into batches in $B$, we can send each batch $B^{(i,j)}$ to a different GPU and continue the forward pass from this point onward on multiple GPUs, or process one batch at a time on a single GPU. This enables us to execute dense prediction tasks with very deep networks and for large images, while always preserving the spatial information.
    }
    \label{fig:split-merge}
\end{figure*}

Most state-of-the-art approaches for dense prediction adapt existing CNNs which were originally designed for image classification.
However, these adapted CNNs lose a vast amount of spatial information due to subsampling. As a result this reduces the prediction performance, mostly because of missing small objects or predicting coarse and inaccurate object boundaries. In order to mitigate this problem, other methods are proposed, such as progressive upsampling \cite{eigen-nips14, eigen-iccv15, long-shelhamer-fcn-2015}, deconvolution (or transpose convolutions) \cite{Noh_2015_ICCV}, skip connections \cite{U-net,Hariharan_2015_CVPR,badrinarayanan2017segnet}, utilizing multiple scales of features \cite{jafari_icra17, zhao2017pspnet}, and attention mechanism \cite{zhao2018psanet}. 
Another line of work preserves the resolution by replacing subsampling layers with dilated convolution layers \cite{ChenPKMY14,YuKoltun2016,Yu2017} and widely used in other approaches \cite{zhao2017pspnet, zhao2018psanet,SunXLW19,WangSCJDZLMTWLX19}. In theory, it is possible to design a dilated convolution network without using any subsampling operation to obtain an output with the same size as the input. Unfortunately, for large input sizes the training and inference of the resulting CNN is extremely slow or even sometimes intractable due to limited amount of memory on a GPU, hence subsampling is still needed in practice.

In this work, we propose novel pooling layers called Split-Merge Pooling (SMP). The split pooling layer splits (re-arranges) a feature tensor along its spatial dimension and treats the resulting splits as individual batches (see Fig.~\ref{fig:split-merge}). The split pooling reduces the spatial size with a fixed scaling factor, related to the size of the non-overlapping pooling window. The merge layer acts as the reverse of split pooling operation, i.e., it receives the split batches and re-arranges the elements of the batches to their original locations. In our experiments, we replace all subsampling layers of standard ResNet networks by our split pooling. Finally, we merge the resulting split batches using merge pooling layers to output full resolution prediction. In this way, we do not lose any spatial information since it is preserved in the split batches.
The batches are processed independently after the split, which has the major advantage that after the split each batch can be processed on a different GPU. This enables us to perform dense prediction tasks with very large networks on large images, while always preserving the spatial information.
Furthermore, to increase training speed, we introduced the Shrink and Expand pooling layers as a batch-subsampling of the split and merge pooling.

\noindent
In summary, our contributions are as follows:
\begin{itemize}
    \item We propose a novel pooling method called Split-Merge Pooling (SMP) which enables the unique mapping of each input element to one output element, without losing any spatial information.
    \item We propose a batch-subsampling variant of SMP, termed Shrink-Expand Pooling, to make the training efficient and tractable for very deep networks.
    \item To show the effectiveness of SMP on dense prediction tasks, we choose semantic segmentation and apply SMP to ResNet networks with varying depths. We chose ResNet as our baseline since it is used as the backbone in state-of-the-art approaches. For semantic segmentation of large images from Cityscapes \cite{Cordts2016Cityscapes} and GTA-5 \cite{Richter_2016_ECCV} datasets, our SMP networks outperform the corresponding ResNet networks by a significant margin, with up to $6.8\%$ in IoU score.
    \item We even observe that a SMP version of a shallow ResNet (ResNet18) outperforms the original ResNet101 by $2.8\%$ in IoU score, although ResNet101 is 3.8 times deeper than ResNet18.
\end{itemize}

\section{Related Work}
Utilizing CNNs for image classification became very popular with the introduction of the Imagenet challenge \cite{imagenet} and after some popular network architectures such as Alexnet \cite{Alexnet}, VGG \cite{VGG} and ResNet \cite{He_2016_CVPR} emerged. 
Pooling layers in CNNs were originally introduced for image classification tasks on MNIST dataset \cite{RanzatoHBL07}. The aim of pooling layers is to summarize the information over a spatial neighborhood. 

\paragraph{Dense prediction tasks.} Most of the dense prediction models are based on adapted versions of image classification networks. Eigen \etal \cite{eigen-nips14, eigen-iccv15} adapt the Alexnet \cite{Alexnet} for single image depth estimation. Since the output of such networks is very coarse, they upsample the output progressively and combine it with local features from the input image. Long \etal \cite{long-shelhamer-fcn-2015} introduce the first fully covolutional network (FCN) for semantic image segmentation by adapting VGG network \cite{VGG} for this task. They map intermediate features with higher resolution to label space and combine them with the coarse prediction of the network in order to recover the missing spatial information. Noh \etal \cite{Noh_2015_ICCV} introduce more parameters in a decoder consisting of transpose convolutions (Deconvolution layers) to upsample the coarse output of the encoder. However, it is still difficult to recover the missing information from the coarse output with this approach. Furthermore, other approaches \cite{U-net, Hariharan_2015_CVPR,badrinarayanan2017segnet} use skip connections between encoder and decoder to obtain more detailed information from lower level features.

Yu \etal introduced the concept of dilated convolutions \cite{YuKoltun2016} and later developed the dilated residual network (DRN) \cite{Yu2017}, which uses dilated convolutions to preserve the spatial information. However, they still use three subsampeling layers to reduce the spatial resolution to make training  feasible, i.e. to fit the model into memory. Therefore, DRNs lose spatial information and need upsampling to obtain the full size output prediction. In contrast, our approach does not lose any spatial information. The batch-based design of our split pooling gives us the flexibility of distributed processing of batches on multiple GPUs or sequential processing of batches on one GPU during inference time (see Sec \ref{sec:split-merge}). Furthermore, for training, we learn our network weights using only one subsampled split batch to make the training faster and tractable (see Sec \ref{sec:shrink-expand}).

\begin{figure*}[!htb]
    \centering
    \includegraphics[width=\linewidth]{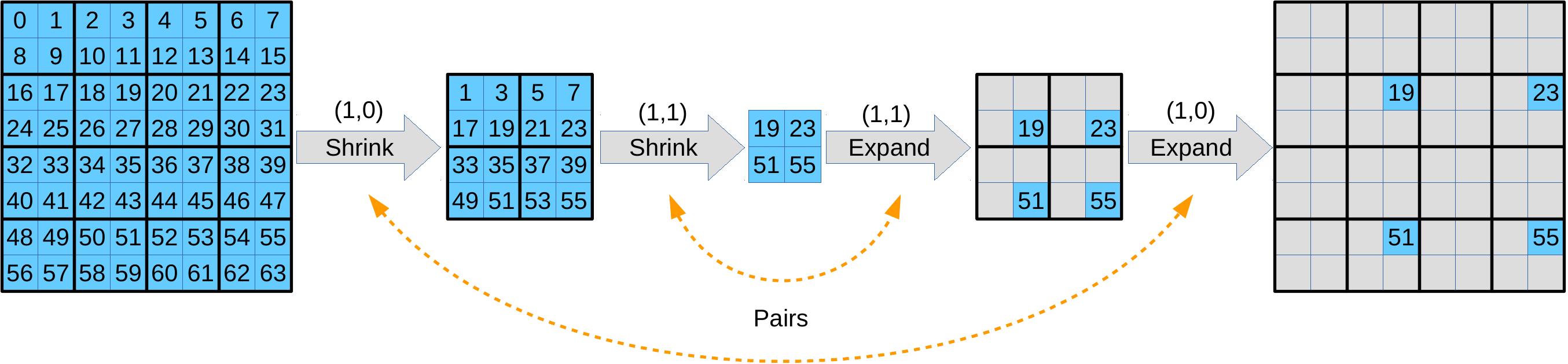}
    \caption{\textbf{Shrink and Expand Pooling example.} The illustration of two pairs of shrink and expand pooling layers with window size of $2\times 2$ and sampling location of $(1,0)$ and $(1,1)$. During training the expand pooling layers back-propagate the error of sparse elements.}
    \label{fig:shrink-expand-seq}
\end{figure*}


\section{Method}
The goal of this work is to design a set of pooling layers that preserve  spatial information, in contrast to subsampling operations such as max-pooling and striding convolutions. Additionally, the new pooling layers have to be applicable to very deep networks. Losing spatial information caused by subsampling is a prevalent problem for dense prediction tasks, such as object detection, semantic segmentation, instance segmentation, or depth estimation. Small objects in the input image can get lost completely with subsampling. Moreover, recovering precise object boundaries in an image becomes challenging due to missing spatial information. These problems, caused by losing spatial information, obviously reduced prediction accuracy. However, simply storing all the information with the original spatial resolution is not a practical option since it causes storage issues, in particular for training on large image datasets. 

Our idea is to preserve all the spatial information by splitting the feature tensor into multiple downsampled batches (see Fig.~\ref{fig:split-merge}) instead of preserving it inside the original spatial resolution. In this way, we have both advantages of large receptive fields due to standard downsampling, as well as keeping all the available spatial information for obtaining precise dense predictions.

\subsection{Split-Merge Pooling}
\label{sec:split-merge}
The split pooling layer splits an input feature tensor along spatial dimensions and outputs each split as a batch. The number of output batches depends on the window size of split pooling, \eg a split pooling with a window size of $2\times 2$ splits the input into 4 batches (see Fig.~\ref{fig:split-merge}). The purpose of the split pooling is to downsample the input while preserving the whole information. Hence, the pooling window covers each element of the input once, without any gap or overlap. 

The merge pooling layer acts as the inverse of split pooling, i.e., it takes the split batches and merges them into one batch. For dense prediction tasks, we apply the same number of merge pooling layers on the final output of network as the number of split pooling layers in a network. The result will be a one-to-one mapping between the input and output pixels of the network.

Formally, given an input $A_{W\times H}$, a split pooling with window size of $w\times h$ splits $A$ into $w*h$ batches 
\[\mathcal{B}=\{\,B^{(k,l)}\mid 0\leq k<w \land 0\leq l<h \,\}.\] 
The elements of $A$ are assigned to batch $B^{(k,l)}$ as follows
\begin{equation}
\label{eq:split}
b^{(k,l)}_{i,j} = a_{i*w+k,j*h+l}
\end{equation}
for all $0\leq i < W/w$ and $0\leq j<H/h$. The spatial size of each batch will be $W/w\times H/h$. If the input to the split pooling consist of multiple batches, the split pooling splits each input batch separately and returns all resulting splits as a set of batches. 

The merge pooling performs the inverse of Eq.~\ref{eq:split}, \ie, if $\mathcal{B}$ is the input and $A$ is the output, the elements of $\mathcal{B}$ are assigned to $A$ as follows
\begin{equation}
\label{eq:merge}
a_{i*w+k,j*h+l} = b^{(k,l)}_{i,j}
\end{equation}
for all $0\leq k<w$, $0\leq l<h$, $0\leq i < W/w$ and $0\leq j<H/h$.

Although preserving the spatial information is beneficial, it increases the number of elements to store during the forward pass. 

The advantage of splitting the inputs into batches is that we can distribute the computation of batches on multiple GPUs or processing them sequentially on one GPU. This is ideal for inference. However, in contrast to inference, training phase additionally requires to compute and store the gradients. 
This makes the training of very deep networks intractable for large batches of inputs with high resolution images.
We handle this issue by introducing Shrink-Expand pooling layers.

\subsection{Shrink-Expand Pooling}
\label{sec:shrink-expand}
The batch-based design of the split pooling layer makes the forward process of each part (split batch) of the input independent of the other parts (split batches). Furthermore, the one-to-one mapping between the elements of input and output gives a clear path between these elements in both forward and backward direction. Giving these two properties, it is possible to train a network using a subset of split batches produced by each split pooling layer. If we reduce the size of the batch subset to one, the space complexity will be the same as the space complexity of the max pooling and striding convolutions. Also, the training time complexity stays the same.

Giving this reasoning, we introduce the shrink and expand pooling layers, which are the batch-subsampled versions of the split and merge pooling layers. The shrink pooling samples one element at a fixed location within the pooling window, and the corresponding expand pooling uses the same fixed location to perform the reverse of the shrink pooling (see Fig.~\ref{fig:shrink-expand-seq}). Hence, the output of expand pooling is sparse. In other words, the shrink pooling is the same as split pooling except it samples only one of the split batches and returns it.

The sampling location $(i,j)$ is set randomly in each forward pass during training to avoid overfitting to part of the training data. For each pair of shrink and expand pooling in the network, we sample only two numbers $(i,j)$. Fig.~\ref{fig:shrink-expand-seq} shows an example of using a sequence of shrink and expand pooling layers. During training, the expand layers only backpropagate the error of sparse valid elements.


\section{Experiments}
We evaluate the effectiveness of our approach for the semantic segmentation task. To examine the impact of our pooling layer, we modify the ResNet and then compare the performance of the modified version and the original one.

\begin{figure}[!htb]
    \centering
    \includegraphics[width=\linewidth]{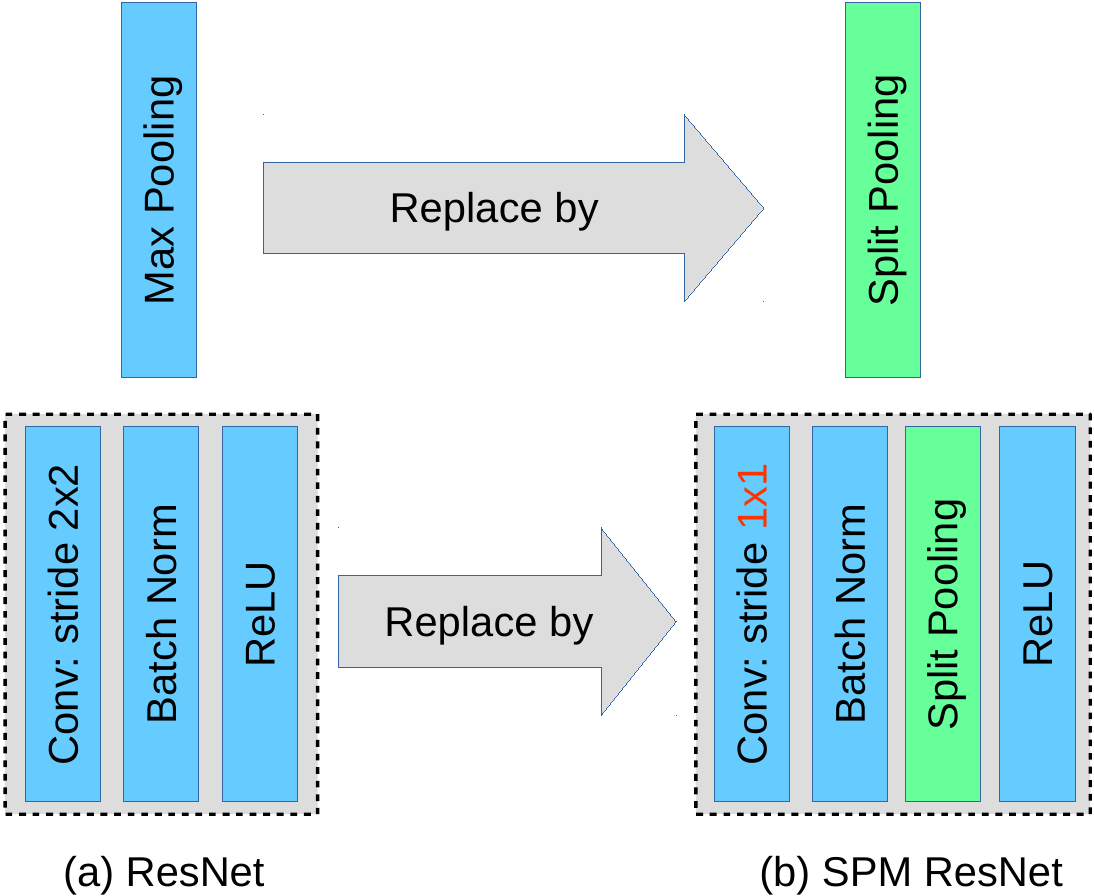}
    \caption{\textbf{Applying split pooling to ResNet.} For applying split pooling to ResNet, we replace the max pooling layer with split pooling (top). Furthermore, we add a split pooling layer after batch normalization layer of convolutional blocks with the stride of $2\times2$ and set the stride to $1\times1$ (bottom).}
    \label{fig:ResNet_adapt}
\end{figure}

\subsection{Experimental Setup}
\paragraph{Baseline FCN32s.} Our experimental models are based on FCN32s \cite{long-shelhamer-fcn-2015} with a variant of ResNet \cite{He_2016_CVPR} as backbone. We adapt the ResNet for semantic segmentation task by removing the average pooling and fully connected layer and replacing them by a $1\times 1$ convolutional layer which maps the output channels of last layer (layer4) to the number of semantic classes. We refer to this model as FCN32s and use it as baseline. FCN32s predictions are 32 times smaller than the input image to the network; thus, the coarse predictions are upsampled to full resolution using bilinear interpolation.

\begin{figure}[!htb]
    \centering
    \begin{tabular}{cc}
    \includegraphics[width=0.23\textwidth]{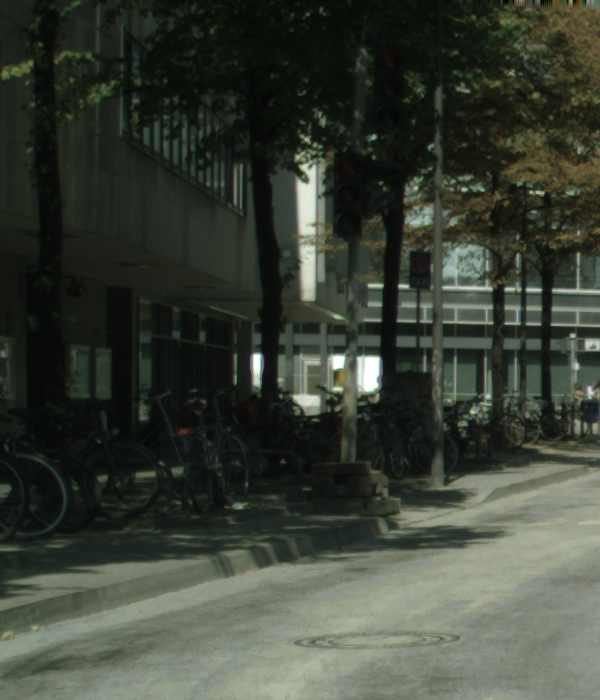} &
    \includegraphics[width=0.23\textwidth]{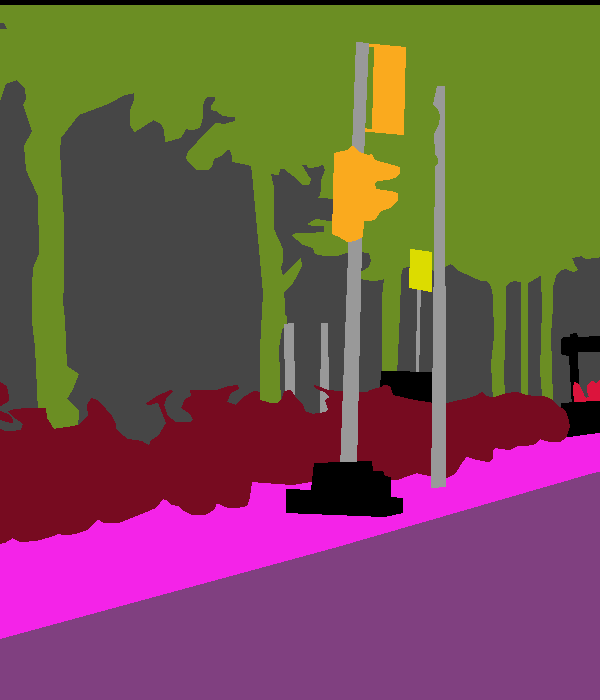} \\
    (a) Image & (b) Ground-truth \\
    \\
    \includegraphics[width=0.23\textwidth]{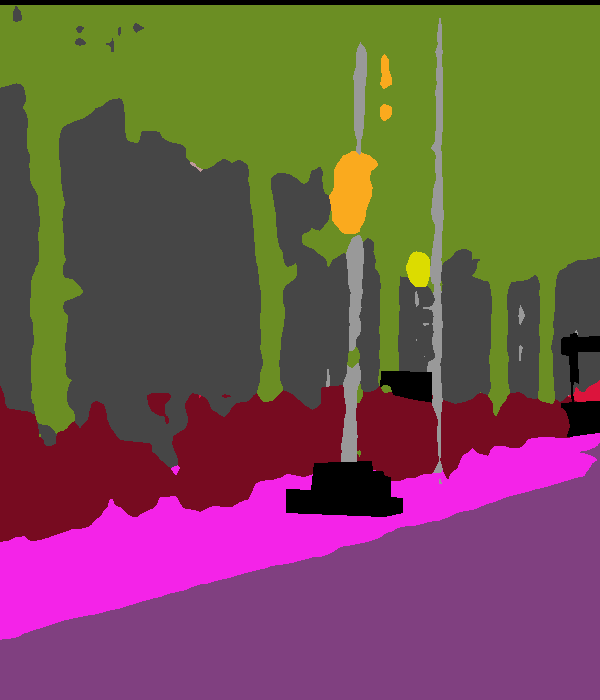} &
    \includegraphics[width=0.23\textwidth]{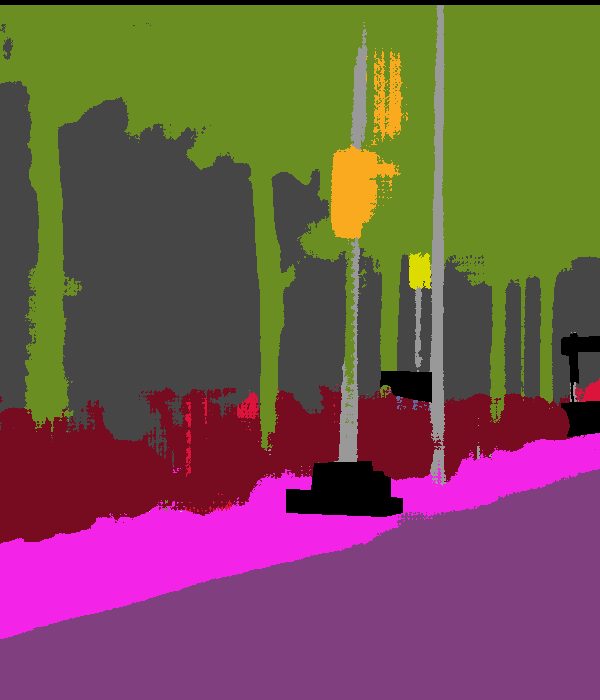}\\
    (c) FCN8s & (d) Ours
    \end{tabular}
    \caption{\textbf{An example of a slightly inaccurate annotation in Cityscapes dataset.} Although the trunk of the tree in the image (a) is visible through bicycles, it is labeled as bicycle in ground truth (b). Our method can obtain detailed boundaries (d) while FCN8s with max pooling cannot (c). Hence, to fully validate the full potential of our method we need a pixel-accurate ground-truth, such as GTA-5.}
    \label{fig:cs-inaccurate}
\end{figure}

\paragraph{SMP-\{18, 34, 101\}.} As illustrated in Fig.~\ref{fig:ResNet_adapt}, in order to apply SMP to ResNet backbone of FCN32s, we simply replace the max pooling layer with a split pooling layer and add a split pooling layer after the batch normalization layer of convolutional blocks (Conv-BN-ReLU) with a stride of $2\times 2$ and set their strides to $1\times 1$. In our experiments, the window size of split pooling layers is $2\times 2$. We call the resulting model, according to the type of backbone ResNet, SMP-18, SMP-34 and SMP-101. The output of SMP-X consist of 1024 batches, since we always have 5 SMP layers each giving 4 batches. We merge the output split batches by performing merge pooling five times. The final result has the same resolution as the input. During training phase, we replace split and merge pooling layer by shrink and expand layers.

\paragraph{FCN8s.} Furthermore, we compare our models to the FCN8s model with original ResNet backbone. The main difference between FCN32s and FCN8s is that FCN8s has two extra $1\times 1$ convolution layers to map the features channel of layer2 and layer3 outputs of ResNet to the number of semantic classes. Then, the output of the network and these new convolutions are resized to the same size and are added together. The final output of FCN8s is 8 times smaller than the input image and should be upsampled to full resolution using bilinear interpolation.

\subsection{Implementation Details}
\paragraph{Data Augmentation.} For all the models we used random crop of size $512\times 512$ and horizontal flip data augmentation. We used a fixed seed for data augmentation for all the experiments.

\paragraph{Training.} We initialize the ResNet backbones with pretrained models from Imagenet. We optimize the parameters using Adam solver \cite{KingmaB14} with learning rate of $1e-5$ and weight decay of $5e-4$. We use the batch size of 10 for all the experiments.

\setlength{\tabcolsep}{4pt}
\begin{table}[!htb]
    \centering
    \begin{tabular}{llccc}
        \hline
         & Backbone &  Pooling & IoU & Size \\
        \hline \hline
        FCN32s & ResNet34 & MP & 64.5 &  21.29M\\ 
        FCN8s  & ResNet34 & MP & 67.8 &  21.30M\\
        \hline
        SMP-34 (ours) & ResNet34 & SMP & \textbf{68.8} & 21.29M\\
        \hline
        \hline
        FCN32s & ResNet101 & MP & 65.5 & 42.54M\\ 
        FCN8s  & ResNet101 & MP & 69.1 & 42.56M\\
        \hline
        SMP-101 (ours) & ResNet101 & SMP & \textbf{69.2} & 42.54M\\
        \hline
    \end{tabular}
    \vspace{0.1cm}
    \caption{\textbf{Cityscapes quantitative results.} Please note that the architecture of SMP-X networks is the same as the FCN32s. The FCN8s architecture has two extra $1\times 1$ convolutions.}
    \label{tab:cs}
\end{table}

\setlength{\tabcolsep}{4pt}
\begin{table}[!htb]
    \centering
    \begin{tabular}{llccc}
        \hline
         & Backbone & Pooling & IoU & Size\\
        \hline \hline
        FCN32s & ResNet18 & MP & 71.1 & 11.186M\\ 
        FCN8s  & ResNet18 & MP & 74.2 & 11.193M\\
        \hline
        SMP-18 (ours) & ResNet18 & SMP & \textbf{77.5} & 11.186M\\
        \hline \hline
        FCN32s & ResNet34 & MP & 73.1 & 21.29M\\ 
        FCN8s  & ResNet34 & MP & 76.7 & 21.30M\\
        \hline
        SMP-34 (ours) & ResNet34 & SMP & \textbf{80.2} & 21.29M\\
        \hline
        \hline
        FCN32s & ResNet101 & MP & 74.6 & 42.54M\\ 
        FCN8s  & ResNet101 & MP & 76.7 & 42.56M\\
        \hline
        SMP-101 (ours) & ResNet101 & SMP & \textbf{80.3} & 42.54M\\
        \hline
    \end{tabular}
    \vspace{0.1cm}
    \caption{\textbf{Quantitative results for GTA-5.} Please note that the architecture of SMP-X networks is the same as the FCN32s, while the FCN8s has two extra $1\times 1$ convolutions.}
    \label{tab:gta}
\end{table}

\setlength{\tabcolsep}{2.5pt}
\begin{table}[!htb]
    \centering
    \begin{tabular}{l ccccccc c}
         & \rot{pole}& \rot{traffic light}& \rot{traffic sign}& \rot{person}& \rot{rider}& \rot{motorcycle}& \rot{bicycle} & IoU  \\
         \hline
         FCN32s-Res34 & 35.4 & 51.5 & 63.0 & 70.4 & 50.3 & 51.8 & 68.4 & 55.8 \\
         FCN8s-Res34 & 53.5 & 57.6 & 69.9 & 77.1 & 53.6 & 51.0 & 72.2 & 62.1 \\
         \hline
         SMP-34(ours) & \textbf{60.5} & \textbf{63.7} & \textbf{74.5} & \textbf{78.8} & \textbf{54.1} & \textbf{52.9} & \textbf{73.9} & \textbf{65.5}  \\
         \hline
         \hline
         FCN32s-Res101 & 39.2 & 58.1 & 66.9 & 71.9 & 51.4 & 53.9 & 70.6 & 58.9 \\
         FCN8s-Res101 & 56.1 & 63.1 & 72.2 & 78.2 & 55.4 & 52.8 & 74.7 & 64.6 \\
         \hline
         SMP-101(ours) & \textbf{63.3} & \textbf{68.7} & \textbf{75.8} & \textbf{79.9} & \textbf{56.1} & \textbf{52.9} & \textbf{76.6} & \textbf{67.6}\\
         \hline
    \end{tabular}
    \vspace{0.1cm}
    \caption{\textbf{Performance on Cityscapes for small and thin objects.}}
    \label{tab:cs-classes}
\end{table}

\setlength{\tabcolsep}{2.5pt}
\begin{table}[!htb]
    \vspace{-0.3cm}
    \centering
    \begin{tabular}{l ccccccccccccccccccc c}
         & \rot{pole}& \rot{traffic light}& \rot{traffic sign}& \rot{person}& \rot{rider}& \rot{motorcycle}& \rot{bicycle} & IoU  \\
         \hline
         FCN32s-Res18 & 44.6 & 45.3 & 60.7 & 69.3 & 64.5 & 59.8 & 38.5 & 54.7\\
         FCN8s-Res18 & 54.0 & 52.7 & 63.8 & 74.4 & \textbf{70.4} & 66.5 & 40.3 & 60.3\\
         \hline
         SMP-18(ours) & \textbf{72.7} & \textbf{69.2} & \textbf{73.6} & \textbf{76.8} & 67.8 & \textbf{69.3} & \textbf{46.5} & \textbf{68.0} \\
         \hline
         \hline
         FCN32s-Res34 & 46.0 & 47.7 & 63.0 & 70.0 & 67.9 & 63.9 & 49.5 & 58.3 \\
         FCN8s-Res34 & 55.6 & 57.2 & 68.3 & 76.3 & \textbf{74.0} & 65.8 & 51.3 & 64.1 \\
         \hline
         SMP-34(ours) & \textbf{74.3} & \textbf{71.4} & \textbf{73.7} & \textbf{81.0} & 70.5 & \textbf{73.7} & \textbf{58.5} & \textbf{71.9}\\
         \hline
         \hline
         FCN32s-Res101 & 47.3 & 50.3 & 68.7 & 70.3 & 63.9 & 66.2 & 58.1 & 60.7\\
         FCN8s-Res101 & 57.1 & 59.0 & 70.8 & 75.5 & 67.1 & \textbf{70.6} & 62.4 & 66.1\\
         \hline
         SMP-101(ours) & \textbf{76.8} & \textbf{74.9} & \textbf{78.9} & \textbf{79.0} & \textbf{69.2} & 69.6 & \textbf{67.8} & \textbf{73.7}\\
         \hline
    \end{tabular}
    \vspace{0.1cm}
    \caption{\textbf{Performance on GTA-5 for small and thin objects}}
    \label{tab:gta-classes}
\end{table}

\subsection{Cityscapes}
The Cityscapes dataset \cite{Cordts2016Cityscapes} consists of images with the size of $2048\times 1024$. The images are annotated with 19 semantic classes. We evaluate the performance of our SMP-34 and SMP-101 on Cityascapes validation set and compare it with FCN32s and FCN8s with ResNet34 and ResNet101 backbone networks. The full size images are used for evaluation, \ie without downsampling or cropping.

\setlength{\tabcolsep}{2pt}
\begin{figure*}[!htp]
    \centering
    \begin{tabular}{lc}
        & \begin{tabularx}{\textwidth}{>{\centering\arraybackslash}X>{\centering\arraybackslash}X>{\centering\arraybackslash}X>{\centering\arraybackslash}X}
        FCN32s & FCN8s & SMP (ours) & \shortstack{Image / \\ Ground Truth}
        \end{tabularx}\\
    \rot{\hspace{0.3cm}ResNet101 \hspace{0.5cm}ResNet34} & \includegraphics[width=0.96\linewidth]{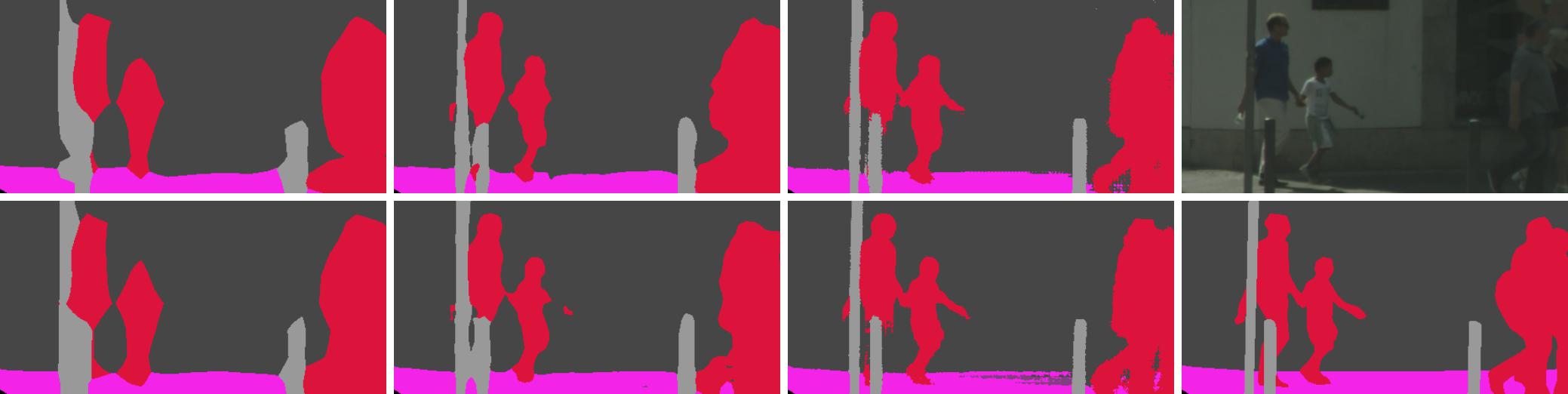} \\
    \\
    \rot{\hspace{0.3cm}ResNet101 \hspace{0.5cm}ResNet34} & \includegraphics[width=0.96\linewidth]{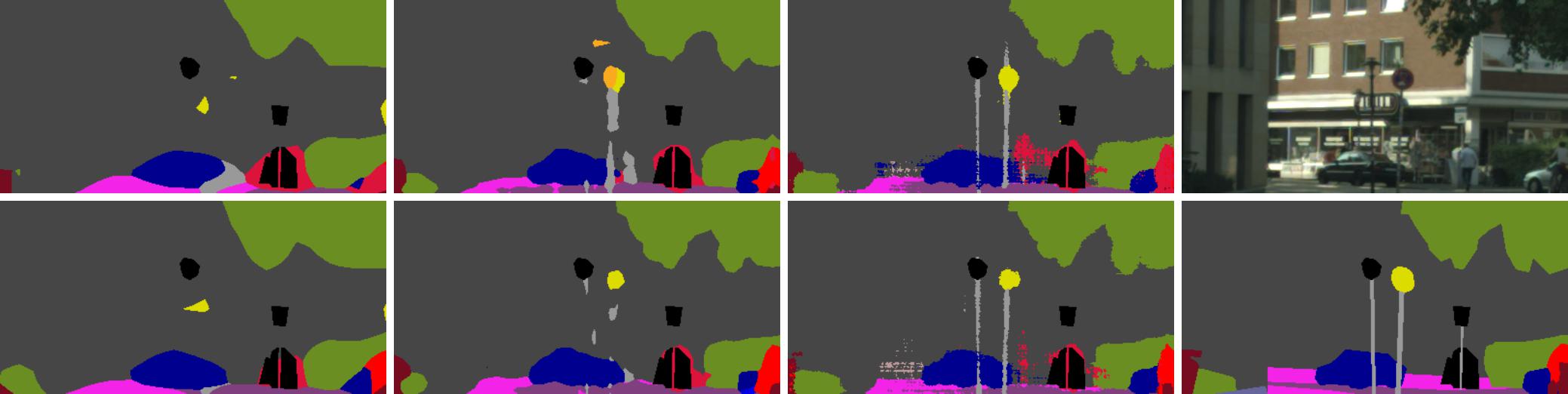} \\
    \\
    \rot{\hspace{0.3cm}ResNet101 \hspace{0.5cm}ResNet34} & \includegraphics[width=0.96\linewidth]{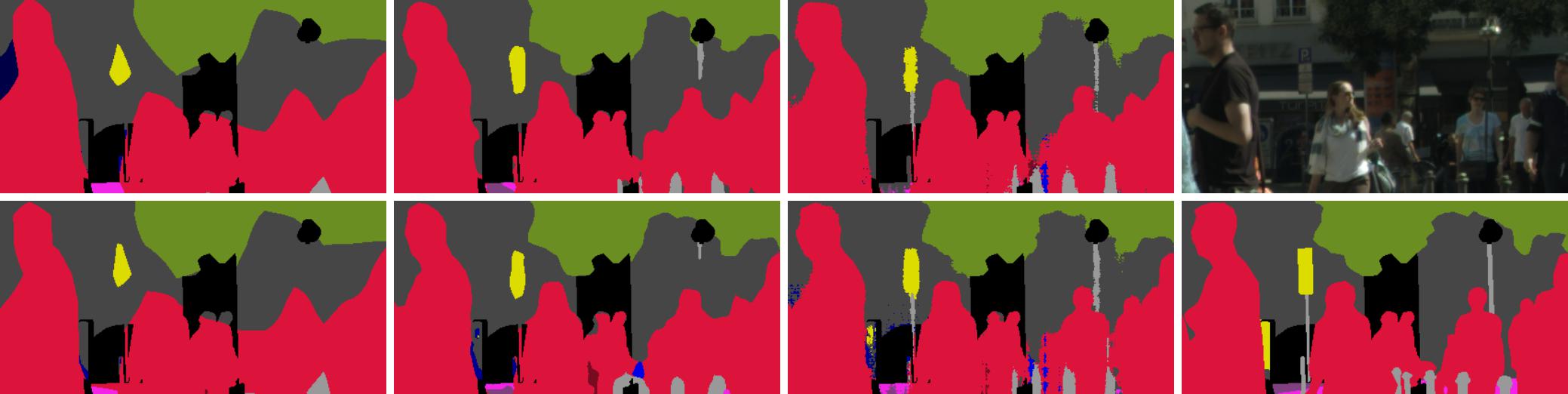}\\
    \\
    \rot{\hspace{0.3cm}ResNet101 \hspace{0.5cm}ResNet34} & \includegraphics[width=0.96\linewidth]{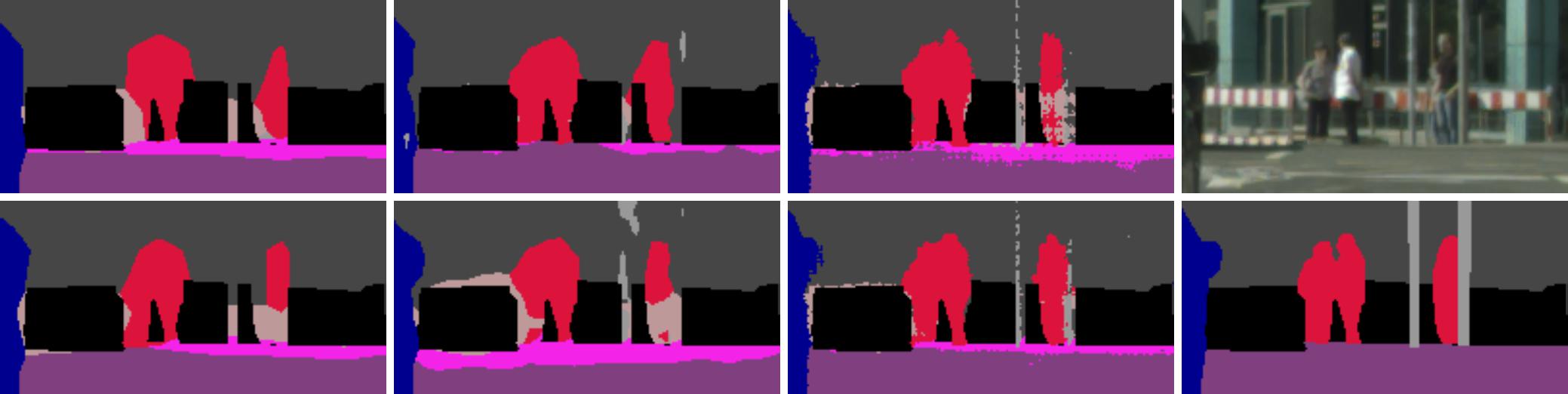} \\
    \end{tabular}
    \caption{\textbf{Cityscapes qualitative results.} In each block, the top row is related to models with ResNet34 backbone and the bottom row to ResNet101. The last column of each block shows the input image (top) and the ground-truth (bottom). To enhance the visual comparison of the results, we have cropped the output labelling. Further results, also in full resolution, can be found in supplementary material.\vspace{1cm}}
    \label{fig:cs-qual}
\end{figure*}

\begin{figure*}[!htp]
    \centering
    \begin{tabular}{cc}
        & \begin{tabularx}{\textwidth}{>{\centering\arraybackslash}X>{\centering\arraybackslash}X>{\centering\arraybackslash}X>{\centering\arraybackslash}X}
        FCN32s & FCN8s & SMP (ours) & \shortstack{Image / \\ Ground Truth}
        \end{tabularx}\\
    \rot{\hspace{0.3cm}ResNet101 \hspace{0.6cm}ResNet34 \hspace{0.7cm}ResNet18} & \includegraphics[width=0.96\linewidth]{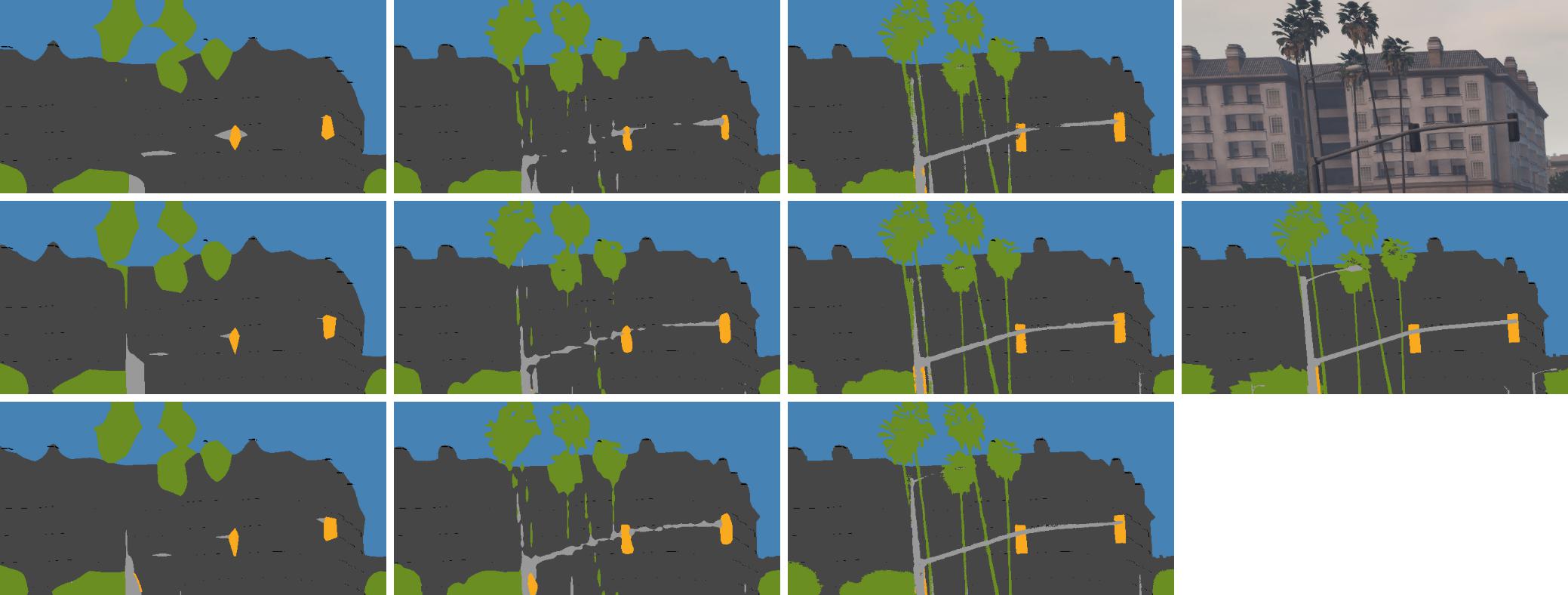} \\
    \\
    \rot{\hspace{0.3cm}ResNet101 \hspace{0.6cm}ResNet34 \hspace{0.7cm}ResNet18} & \includegraphics[width=0.96\linewidth]{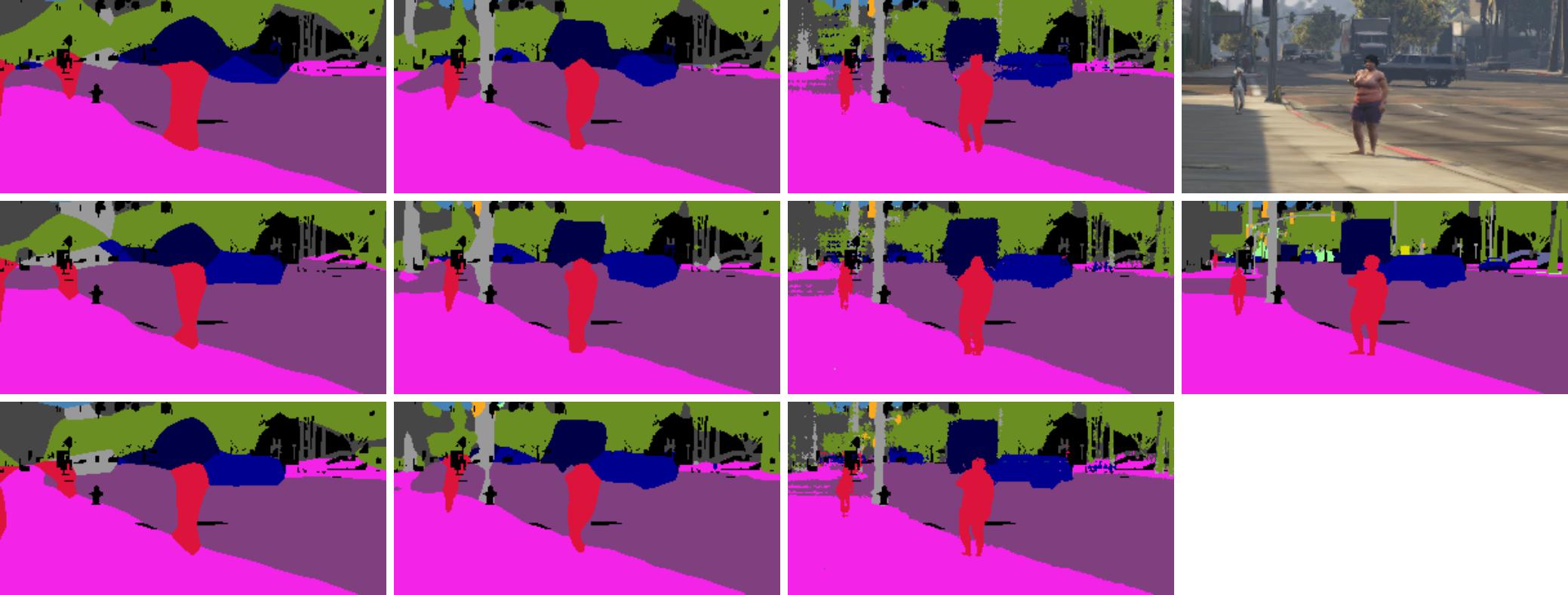} \\
    \\
    \rot{\hspace{0.3cm}ResNet101 \hspace{0.6cm}ResNet34 \hspace{0.7cm}ResNet18} & \includegraphics[width=0.96\linewidth]{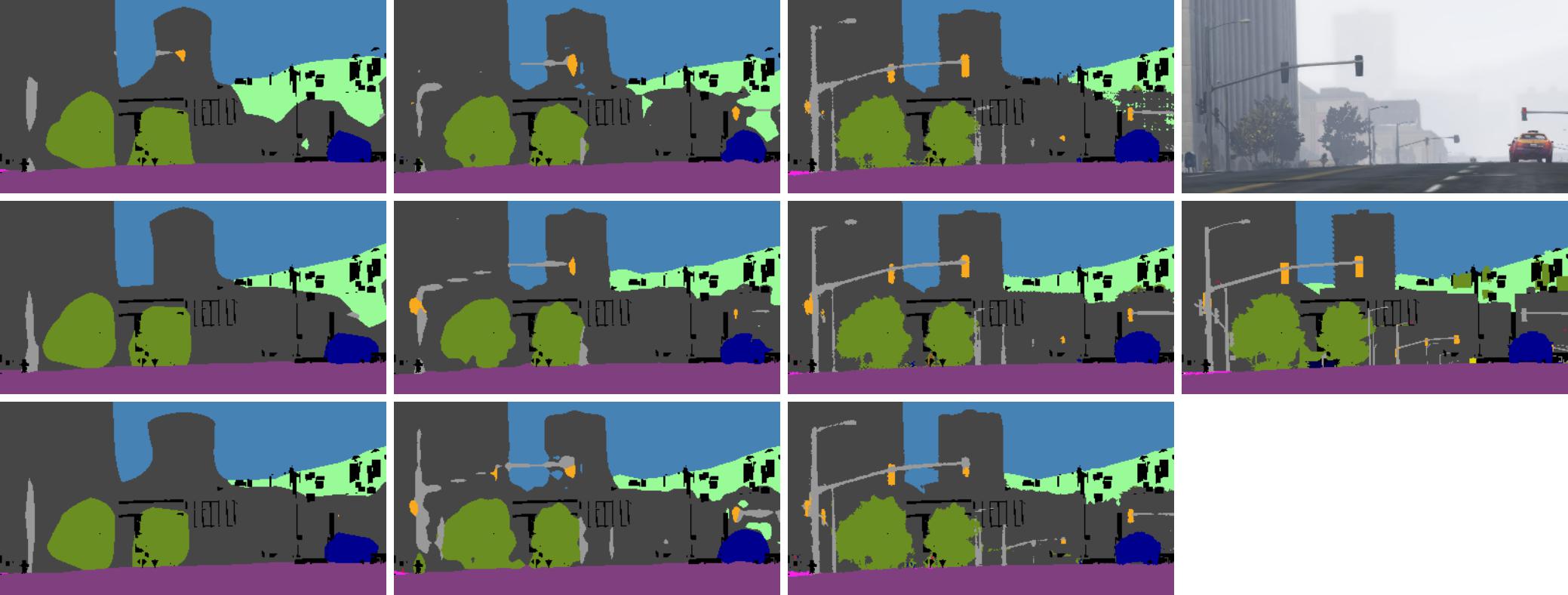}
    \end{tabular}
    \caption{\textbf{GTA-5 qualitative results.} In each block, the top row is related to ResNet18 backbone, middle row to ResNet34, and bottom row to ResNet101. The last column of each block shows the input image (top) and the ground-truth (bottom). To enhance the visual comparison of the results, we have cropped the output labelling. Further results, also in full resolution, can be found in supplementary material.}
    \label{fig:gta-qual}
\end{figure*}

Table~\ref{tab:cs} summarizes the comparison of different methods with respect to their setups, performances and number of parameters (size). Each SMP model outperforms the corresponding FCN32s and FCN8s with the same ResNet backbone significantly. SMP-34 even outperforms FCN32s-ResNet101 by $3.3\%$, although the latter is almost 2 times larger. 
The performance of our approach on objects from small and thin classes is reported in Table~\ref{tab:cs-classes}. As we can see in the table, SMP models outperforms their corresponding original models on small and thin objects. Both SMP models outperform their corresponding FCN32s models for \textit{pole} by $24\%$, for \textit{traffic light} by more than $10\%$, for \textit{traffic sign} by more than $9\%$, and for  \textit{person} by $8\%$. As it is shown in Fig.~\ref{fig:cs-qual}, the improvement of these classes can be noticed visually as well. The performance for the remaining objects is mostly better, or sometimes slightly worse. 

Qualitative results are shown in Fig.~\ref{fig:cs-qual}. For the sake of improved visibility we have cropped the results. The full size output images can be found in supplementary material.

\subsection{GTA-5}

Cityscapes is one of the most accurately annotated semantic segmentation datasets, however it is still not pixel-accurate (see Fig.~\ref{fig:cs-inaccurate}). Obtaining pixel-accurate annotations from real data is extremely challenging and expensive. Therefore, for analysing the full potential of our method, we evaluate our method on GTA-5 \cite{Richter_2016_ECCV}, which is a synthetic dataset with the same semantic classes as Cityscapes. Since GTA-5 is a synthetic, the annotations are pixel-accurate and ideal for our purpose.
The GTA-5 dataset \cite{Richter_2016_ECCV} consist of 24,999 realistic synthetic images with pixel-accurate semantic annotations. We randomly select 500 images as validation set, which we did not use for training. 

Table~\ref{tab:cs} summarizes the comparison of different methods with respect to their setups, performances and number of parameters (size). As we can see, due to the pixel-accurate annotations of GTA-5 dataset, the improvement of our proposed models, over their baselines, is more significant compared to Cityscapes. Each SMP model outperforms the corresponding FCN32s and FCN8s with the same ResNet backbone significantly. Particularly, our SMP-18 even outperforms its FCN32s-ResNet101 and FCN8s-ResNet101 counterparts, although it has 4 times fewer parameters. 
The performance of our approach on objects from small and thin classes is reported in Table~\ref{tab:gta-classes}. As we can see, similarly to Cityscapes, SMP models outperforms their corresponding original models on small and thin objects. Compared to FCN32s models, our corresponding SMP models improve the categories \textit{pole} by more than $28\%$, \textit{traffic light} by more than $23\%$, \textit{traffic sign} by more than $10\%$, and \textit{person} by more than $7\%$. The improvement over these classes is also visually significant (see Fig.~\ref{fig:gta-qual}). 

\subsection{Run-time Analysis}
\begin{figure}[!htb]
	\centering
	\includegraphics[width=\linewidth]{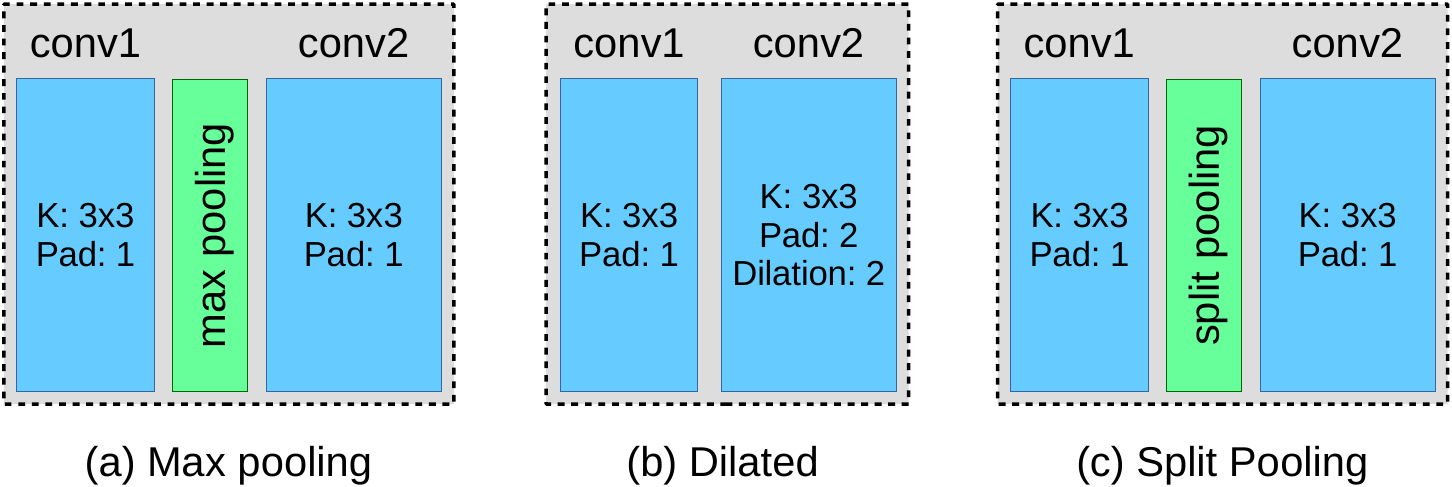}
	\caption[Runtime analysis setup]{Architectures used for runtime analysis. \textit{conv2} in (c) is identical to \textit{conv2} in (a) while \textit{conv2} in (c) is dilated convolution with padding $2$.}
	\label{fig:run-time-archs}
\end{figure}

For analyzing the time complexity of the Split-Merge pooling, we designed small networks to just focus on the proposed pooling layers instead of analyzing the time complexity of them on a particular task with specific network architecture. As it is shown in Fig.~\ref{fig:run-time-archs}, we consider three networks with (a) max pooling, (b) dilated convolution, and (c) Split pooling. 
We choose to compare our proposed pooling setup (c) with dilated convolutions (b) due to the success of the dilated convolution in dense prediction tasks. Almost all state-of-the-art approaches in dense prediction tasks (such as semantic segmentation, depth estimation, optical flow estimation) are using dilated convolutions in their architectures to achieve a detailed output.

In Table~\ref{tab:run-time}, we show the Giga floating-point operation (GFLOP) of each component of each setup for an input tensor of size $1\times3\times256\times256$. As we can see, the dilated convolution setup and split pooling setup have the same GFLOPs which means dilated convolution layers can be replaced with our proposed pooling layers in an arbitrary architecture without changing the complexity of the network. However, our split pooling layer has two advantages: 
\begin{enumerate}
	\item faster training time using shrink-expand layers
	\item faster inference time by parallelizing the forward-computation of split layer output batches (in this example setup computation of \textit{conv2}, see Table~\ref{tab:run-time})
\end{enumerate}

\begin{table}[!htb]
	\centering
	\begin{tabular}{l|cc|c|cc||c}
		& batches & conv1 & pooling & batches & conv2 & total \\
		\hline
		(a) Max Pooling   & 1 & 0.23 & 0 & 1 & 2.42 & 2.65 \\
		(b) Dilated Conv. & 1 & 0.23 & - & 1 & 9.68 & 9.92 \\
		(c) Split Pooling & 1 & 0.23 & 0 & 4 & 9.68 & 9.92 \\
		\hline
	\end{tabular}
	\vspace{0.1cm}
	\caption[Runtime analysis]{GFLOPs of the models calculated on the input size of $1\times3\times256\times256$. Note that the number of batches are increased after split pooling.}
	\label{tab:run-time}
\end{table}

\vspace{-0.2cm}
\section{Conclusion}
We proposed a novel pooling method SMP with the goal of preserving the spatial information throughout the entire network. SMP can be used instead of any subsampling operations in a network architecture. We show that by replacing subsampling operations with SMP in ResNet, we achieved two important properties for any dense prediction task at the same time: i) the network has a large receptive field, ii) the network provides a unique mapping from input pixels to output pixels. Furthermore, the computation of a network with SMP can be distributed to multiple GPUs due to batch-based design of SMP.
We show experimentally that the resulting network outperforms the original one significantly. 

\paragraph{}
\textbf{Acknowledgements} This work is funded by the DFG grant “COVMAP: Intelligente Karten mittels gemeinsamer GPS- und Videodatenanalyse” (RO 4804/2-1).

{\small
\bibliographystyle{ieee_fullname}
\bibliography{egbib}
}

\section*{Appendix A: Detailed Qualitative Results}
Table~\ref{tab:cs-classes} and Table~\ref{tab:gta-classes} show the detailed quantitative results of our proposed models on Cityscapes and GTA-5 datasets respectively.

 \setlength{\tabcolsep}{2.5pt}
 \begin{table*}[!htb]
     \centering
     \begin{tabular}{l ccccccccccccccccccc c}
          & \rot{road}& \rot{sidewalk}& \rot{building}& \rot{wall}& \rot{fence}& \rot{pole}& \rot{traffic light}& \rot{traffic sign}& \rot{vegetation}& \rot{terrain}& \rot{sky}& \rot{person}& \rot{rider}& \rot{car}& \rot{truck}& \rot{bus}& \rot{train}& \rot{motorcycle}& \rot{bicycle} & IoU  \\
          \hline
          FCN32s-Res34 & 97.2 & 78.5 & 88.5 & 40.1 & 48.5 & 35.4 & 51.5 & 63.0 & 88.9 & 56.5 & 89.3 & 70.4 & 50.3 & 91.2 & 48.3 & 64.9 & 43.5 & 51.8 & 68.4 & 64.5 \\
          FCN8s-Res34 & 97.3 & 80.0 & 89.6 & 37.2 & 49.5 & 53.5 & 57.6 & 69.9 & 90.9 & 57.7 & 92.4 & 77.1 & 53.6 & 92.5 & \textbf{49.9} & 69.4 & \textbf{46.6} & 51.0 & 72.2 & 67.8 \\
          \hline
          SMP-34(ours) & \textbf{97.3} & \textbf{80.6} & \textbf{90.5} & \textbf{42.2} & \textbf{50.6} & \textbf{60.5} & \textbf{63.7} & \textbf{74.5} & \textbf{91.4} & \textbf{58.7} & \textbf{92.8} & \textbf{78.8} & \textbf{54.1} & \textbf{92.9} & 48.5 & \textbf{70.6} & 32.3 & \textbf{52.9} & \textbf{73.9} & \textbf{68.8}  \\
          \hline
          \hline
          FCN32s-Res101 & 97.3 & 79.6 & 88.9 & 38.3 & 51.3 & 39.2 & 58.1 & 66.9 & 89.4 & 55.4 & 91.3 & 71.9 & 51.4 & 91.8 & 43.0 & 65.1 & 41.3 & 53.9 & 70.6 & 65.5 \\
          FCN8s-Res101 & 97.5 & 81.2 & 90.3 & \textbf{41.0} & \textbf{49.9} & 56.1 & 63.1 & 72.2 & 91.4 & 59.8 & 92.8 & 78.2 & 55.4 & 92.9 & \textbf{49.9} & \textbf{68.6} & \textbf{44.7} & 52.8 & 74.7 & 69.1\\
          \hline
          SMP-101(ours) & \textbf{97.5} & \textbf{82.7} & \textbf{90.6} & 39.6 & 48.4 & \textbf{63.3} & \textbf{68.7} & \textbf{75.8} & \textbf{91.9} & \textbf{61.0} & \textbf{93.4} & \textbf{79.9} & \textbf{56.1} & \textbf{93.2} & 41.5 & 61.2 & 40.0 & \textbf{52.9} & \textbf{76.6} & \textbf{69.2}\\
          \hline
     \end{tabular}
     \vspace{0.1cm}
     \caption{\textbf{Cityscapes - detailed.} SMP models outperforms their corresponding original models on small and thin objects. Both SMP models outperform their corresponding FCN32s models for \textit{pole} by $24\%$, for \textit{traffic light} by more than $10\%$, for \textit{traffic sign} by more than $10\%$, and for \textit{person} by $8\%$.}
     \label{tab:cs-classes}
 \end{table*}

 \setlength{\tabcolsep}{2.5pt}
 \begin{table*}[!htb]
     \vspace{-0.3cm}
     \centering
     \begin{tabular}{l ccccccccccccccccccc c}
          & \rot{road}& \rot{sidewalk}& \rot{building}& \rot{wall}& \rot{fence}& \rot{pole}& \rot{traffic light}& \rot{traffic sign}& \rot{vegetation}& \rot{terrain}& \rot{sky}& \rot{person}& \rot{rider}& \rot{car}& \rot{truck}& \rot{bus}& \rot{train}& \rot{motorcycle}& \rot{bicycle} & IoU  \\
          \hline
          FCN32s-Res18 & 95.4 & 81.8 & 87.3 & 62.9 & 54.8 & 44.6 & 45.3 & 60.7 & 79.9 & 70.0 & 93.3 & 69.3 & 64.5 & 88.3 & \textbf{83.9} & 87.8 & 82.6 & 59.8 & 38.5 & 71.1\\
          FCN8s-Res18 & 96.1 & 84.6 & 88.6 & 66.6 & \textbf{56.4} & 54.0 & 52.7 & 63.8 & 83.3 & 72.5 & 94.8 & 74.4 & \textbf{70.4} & 89.5 & 80.7 & \textbf{90.4} & 85.1 & 66.5 & 40.3 & 74.2\\
          \hline
          SMP-18(ours) & \textbf{96.5} & \textbf{85.6} & \textbf{89.6} & \textbf{67.2} & 55.7 & \textbf{72.7} & \textbf{69.2} & \textbf{73.6} & \textbf{88.4} & \textbf{75.0} & \textbf{97.9} & \textbf{76.8} & 67.8 & \textbf{90.6} & 82.3 & 80.8 & \textbf{86.1} & \textbf{69.3} & \textbf{46.5} & \textbf{77.5} \\
          \hline
          \hline
          FCN32s-Res34 & 96.5 & 85.2 & 88.0 & 64.2 & 55.4 & 46.0 & 47.7 & 63.0 & 80.9 & 72.4 & 93.5 & 70.0 & 67.9 & 89.2 & 86.2 & 85.1 & 84.2 & 63.9 & 49.5 & 73.1 \\
          FCN8s-Res34 & 97.0 & 87.6 & 89.5 & 68.9 & \textbf{59.4} & 55.6 & 57.2 & 68.3 & 83.9 & 74.4 & 94.9 & 76.3 & \textbf{74.0} & 90.5 & 86.8 & \textbf{90.4} & 84.9 & 65.8 & 51.3 & 76.7 \\
          \hline
          SMP-34(ours) & \textbf{97.3} & \textbf{88.1} & \textbf{91.4} & \textbf{69.6} & 58.3 & \textbf{74.3} & \textbf{71.4} & \textbf{73.7} & \textbf{89.1} & \textbf{77.1} & \textbf{98.2} & \textbf{81.0} & 70.5 & \textbf{92.6} & \textbf{88.3} & 83.7 & \textbf{87.6} & \textbf{73.7} & \textbf{58.5} & \textbf{80.2}\\
          \hline
          \hline
          FCN32s-Res101 & 96.6 & 86.0 & 89.2 & 70.4 & 59.3 & 47.3 & 50.3 & 68.7 & 81.6 & 72.6 & 93.7 & 70.3 & 63.9 & 89.6 & 88.2 & \textbf{89.1} & 77.1 & 66.2 & 58.1 & 74.6\\
          FCN8s-Res101 & 96.6 & 86.6 & 89.2 & 61.7 & 60.6 & 57.1 & 59.0 & 70.8 & 84.5 & 73.9 & 95.1 & 75.5 & 67.1 & 90.0 & \textbf{88.6} & 86.0 & 81.9 & \textbf{70.6} & 62.4 & 76.7\\
          \hline
          SMP-101(ours) & \textbf{97.3} & \textbf{88.7} & \textbf{91.7} & \textbf{71.0} & \textbf{62.2} & \textbf{76.8} & \textbf{74.9} & \textbf{78.9} & \textbf{89.6} & \textbf{79.0} & \textbf{97.8} & \textbf{79.0} & \textbf{69.2} & \textbf{90.6} & 86.0 & 71.9 & \textbf{84.3} & 69.6 & \textbf{67.8} & \textbf{80.3}\\
          \hline
     \end{tabular}
     \vspace{0.1cm}
     \caption{\textbf{GTA-5 - detailed.} SMP models outperforms their corresponding original models on small and thin objects for GTA-5 as well. Compare to FCN32s models, our corresponding SMP models improve the categories \textit{pole} by more than $28\%$, \textit{traffic light} by more than $23\%$, \textit{traffic sign} by more than $10\%$, and \textit{person} by more than $7\%$.}
     \label{tab:gta-classes}
 \end{table*}

\end{document}